\documentclass[10pt,twocolumn,letterpaper]{article}

\usepackage{wacv}
\usepackage{times}
\usepackage{epsfig}
\usepackage{graphicx}
\usepackage{amsmath}
\usepackage{amssymb}
\usepackage{booktabs}

\usepackage{xcolor}
\usepackage{arydshln}
\usepackage{enumitem}
\usepackage{caption}
\usepackage{subfigure}
\usepackage{multirow}
\usepackage{hhline}
\usepackage{nicematrix}
\usepackage{mathabx}
\usepackage[flushleft]{threeparttable}
\usepackage{pgfplots}
\usepackage{tikz}
\usepackage{adjustbox}
\usetikzlibrary{calc,positioning}
\definecolor{lightblue}{RGB}{187, 227, 252}
\definecolor{lightgray}{RGB}{199, 199, 199}
\definecolor{lightyellow}{RGB}{247, 248, 208}
\usepackage{adjustbox}
\usepackage{tablefootnote}
\usepackage[flushleft]{threeparttable}
\usetikzlibrary{spy,arrows}
\usepackage[accsupp]{axessibility}  

\def\wacvPaperID{2} 

\wacvfinalcopy 

\ifwacvfinal
\usepackage[breaklinks=true,bookmarks=false]{hyperref}
\else
\usepackage[pagebackref=true,breaklinks=true,colorlinks,bookmarks=false]{hyperref}
\fi
\usepackage[capitalize]{cleveref}
\crefname{section}{Sec.}{Secs.}
\Crefname{section}{Section}{Sections}
\Crefname{table}{Table}{Tables}
\crefname{table}{Tab.}{Tabs.}
\begin{document}

\title{Learning Pairwise Interaction for Generalizable DeepFake Detection}

\author{Ying Xu$^{1}$, Kiran Raja$^{1}$, Luisa Verdoliva$^{2}$, Marius Pedersen$^{1}$\\
$^1$ Norwegian University of Science and Technology, Norway \\ {\tt\small \{ying.xu, kiran.raja, marius.pedersen\} @ntnu.no}\\
$^2$ University Federico II of Naples, Italy {\tt\small verdoliv@unina.it}
}

\maketitle
\thispagestyle{empty}

\begin{abstract}
A fast-paced development of DeepFake generation techniques challenge the detection schemes designed for known type DeepFakes. A reliable Deepfake detection approach must be agnostic to generation types, which can present diverse quality and appearance. Limited generalizability across different generation schemes will restrict the wide-scale deployment of detectors if they fail to handle unseen attacks in an open set scenario. We propose a new approach, Multi-Channel Xception Attention Pairwise Interaction (MCX-API), that exploits the power of pairwise learning and complementary information from different color space representations in a fine-grained manner. We first validate our idea on a publicly available dataset in a intra-class setting (closed set) with four different Deepfake schemes. Further, we report all the results using balanced-open-set-classification (BOSC) accuracy in an inter-class setting (open-set) using three public datasets. Our experiments indicate that our proposed method can generalize better than the state-of-the-art Deepfakes detectors. We obtain 98.48\% BOSC accuracy on the FF++ dataset and 90.87\% BOSC accuracy on the CelebDF dataset suggesting a promising direction for generalization of DeepFake detection. We further utilize t-SNE and attention maps to interpret and visualize the decision-making process of our proposed network. \url{https://github.com/xuyingzhongguo/MCX-API}
\end{abstract}

\section{Introduction}
\begin{figure}[t!]
    \centering
    \includegraphics[width=0.8\linewidth]{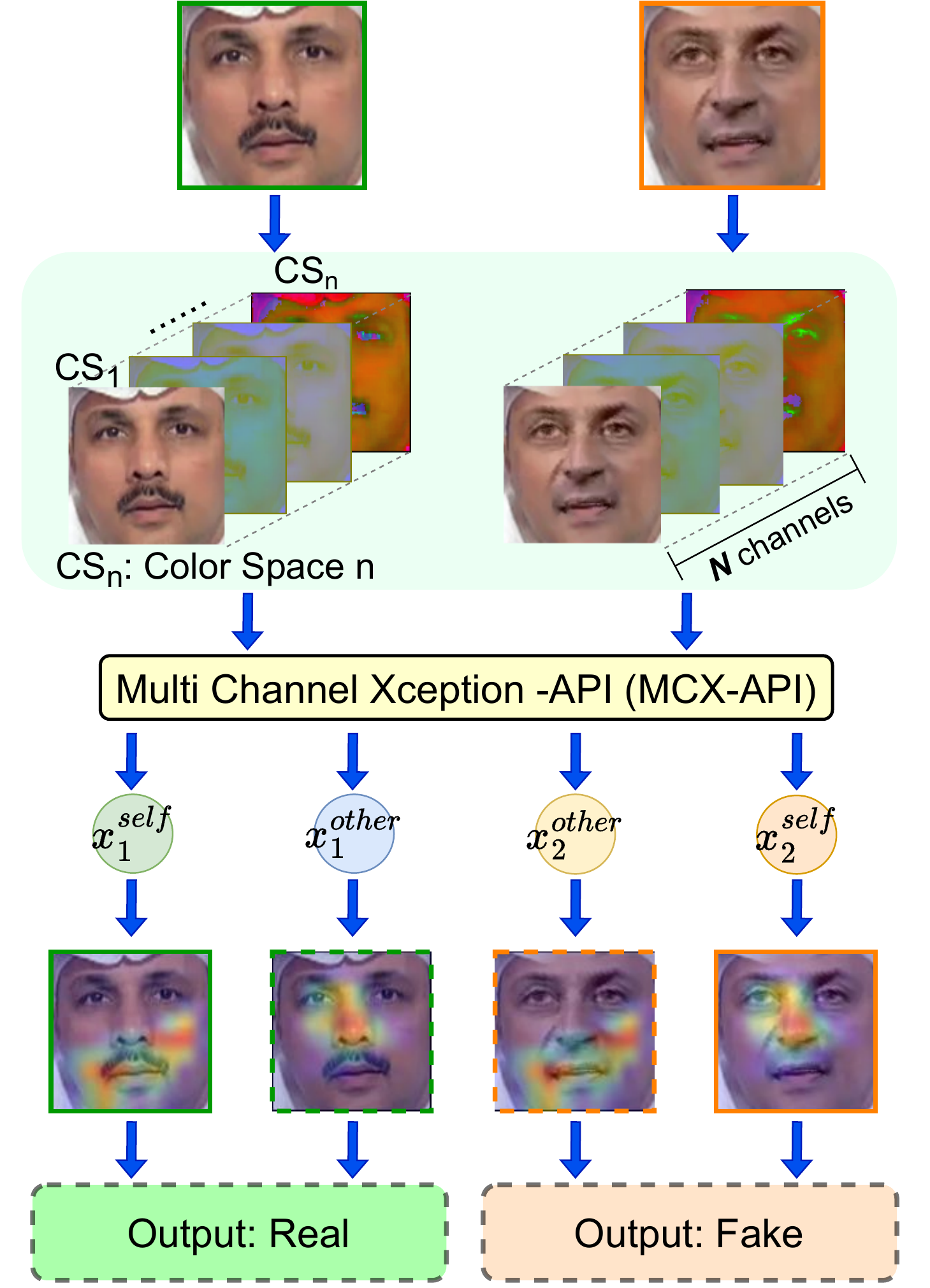}
    \caption{Overview of proposed Multi Channel Xception Attentive Pairwise Interaction (MCX-API) network. Two inputs are first represented in $n$ kinds of color spaces, $CS_1$ to $CS_n$ to obtain a two N-channel input and subsequently feature vectors. We thereafter obtain $x_{i}^{self}$ and $x_{i}^{other}$ by comparison through MCX-API, where $x_{i}^{self}$ is enhanced by its own images and $x_{i}^{other}$ is activated by the other image. $x_{i}$ is therefore improved with discriminative clues that come from both images. By comparison, we can finally distinguish if an image is pristine or fake.}
    \label{fig:overview}
\end{figure}
Deepfakes are synthetic media that are generated by deep learning methods to manipulate the content in images and videos. The manipulations include altering people's identities, faces, expressions, speech or bodies to both entertainment and malicious intent (for example pornographic uses). Benefiting from the remarkable advancement in generation models, amateurish individuals are capable of creating Deepfakes using off-the-shelf models~\cite{DeepFaceLab, fffs, FaceApp} without tedious efforts. 
In the meantime, channelized efforts have been dedicated to devising Deepfakes detection algorithms using multiple approaches such as by determining unique artifacts~\cite{matern2019exploiting,ciftci2020fakecatcher,fernandes2019predicting,haliassos2021lips,agarwal2019protecting,li2020face}, utilizing Convolutional Neural Networks (CNNs) based networks~\cite{marra2019gans,rossler2019faceforensics++,nguyen2019use}, employing frequency domain information~\cite{durall2019unmasking,chen2021attentive,qian2020thinking} and other clues~\cite{cozzolino2021id, cozzolino2022audio}.
\par With an atomic effort, these methods could perform well with an average of more than 99\%~\cite{rossler2019faceforensics++} accuracy in a closed-set problem where the training and testing data are pulled from the same label and feature spaces. For example, the network is trained on attacks A, B and C and tested on images/videos drawn from attack A or B or C. However, newer DeepFakes generation mechanisms make the detection algorithms untrustworthy and non-generalizable by degrading the performance of the detector~\cite{zhao2021learning,zhou2021joint} as no exception to those classifiers trained with machine learning methods. In the context of DeepFakes detection, this can be parallel to detecting attack D when the detector is trained on A, B, and C, making it an open-set problem. The reasons behind the collapse of detection models towards unseen contents can, to some degree, be attributed to various generation algorithms, which often result in different data distributions, feature spaces, and appearance properties of images or videos. While one can see the imperative need for a generalizable detection technique to make reliable decisions on unknown/unseen generation types in addition to known/seen generation data, we note low performances of networks in this direction \cite{zhao2021learning,zhou2021joint,xu2022supervised, aneja2020generalized}. 
\par We thus motivate our work, focusing on both closed-set and open-set detection in this article. We draw our inspiration from how humans tend to detect altered media in a fine-grained manner by comparing one kind of visual content to another. Human decision making relies on detecting an unseen kind of manipulated images/videos as fake by comparing the unknown generation type to the known generation types, especially the artifacts and clues~\cite{zhuang2020learning}. Initial work using on pairwise interaction has shown promising directions to capture subtle differences in a pairwise manner with not only principal parts of the image but also distinct details from the other image \cite{zhuang2020learning}. Using such a paradigm, we propose to learn the known type of generations in a fine-grained pairwise manner explicitly to improve the performance of a Deepfake detector for unknown types. Further, we also note the complementary information an image/video can exhibit in different color spaces along the same lines. We therefore incorporate information from four color spaces, including RGB, CIELab, HSV, and YCbCr integrating to boost the attentive pairwise learning to guide the detector to classify the non-manipulated images efficiently. Our proposed approach exploits the information from color channels in a pairwise manner using the strengths of the Xception network and we refer to this as the Multi-Channel Xception Attentive Pairwise Interaction (MCX-API) network between non-manipulated images against a set of manipulated images and to try to generalize the detector towards unknown manipulation types or unseen data. \Cref{fig:overview} shows an overview of the idea presented in this work. 
\par To validate our idea in this work, we conduct various experiments using FaceForensics++ dataset~\cite{rossler2019faceforensics++} which consists of four different manipulation classes including DeepFakes (DF)~\cite{ffdf}, FaceSwap (FS)~\cite{fffs}, Face2Face (F2F)~\cite{thies2016face2face} and NeuralTextures (NT)\cite{thies2019deferred} where we obtain better state-of-the-art (SOTA) performance or at par detection performance to best performing SOTA approaches in closed-set experiments~\cite{chollet2017xception,afchar2018mesonet,zhao2021learning,li2020face}. Furthermore, we demonstrate the effectiveness of variants of the proposed approach in detecting Deepfakes in open-set scenarios where our approach achieves better results than SOTA models on three other public datasets such as FakeAV~\cite{khalid2021fakeavceleb}, KoDF~\cite{kwon2021kodf}, and Celeb-DF~\cite{li2020celeb}.
\par A detailed ablation study is presented on MCX-API to illustrate the variability of performance of the detector to various design choices in the network. Thus, the main contributions of our paper are \textbf{(1)} We propose a new framework - Multi-Channel Xception Attentive Pairwise Interaction (MCX-API)  for Deepfakes detection by exploiting color space and pairwise interaction simultaneously, bringing a novel fine-grained idea for the Deepfakes detection field. \textbf{(2)}We report all results by balanced-open-set-classification (BOSC) accuracy to exemplify the generalizability of our proposed approach. 
\textbf{(3)}We conduct cross-datasets validations with three SOTA Deepfake datasets, Celeb-DF~\cite{li2020celeb}, KoDF~\cite{kwon2021kodf} and FakeAVCelebDF~\cite{khalid2021fakeavceleb}. Furthermore, we compared the results with SOTA Deepfake detection methods. Our MCX-API obtains 98.48\% BOSC accuracy on the FF++ dataset and 90.87\% BOSC accuracy on the Celeb-DF dataset, indicating an optimistic direction for the generalization of DeepFake detection.

In the rest of the paper, we list a set of directly related works in \cref{sec:related-works} and then present our proposed approach in \cref{sec:proposed-approach}. 
We provide an analysis of explainability in \cref{sec:ExplainableAnalysisofMCX-API} with the set of experiments and results on generalizability detailed in \cref{sec:Resutls}. We finally conclude the work in \cref{sec:conclusion}.

\begin{figure*}[htp]
    \centering
    \includegraphics[width=0.9\linewidth]{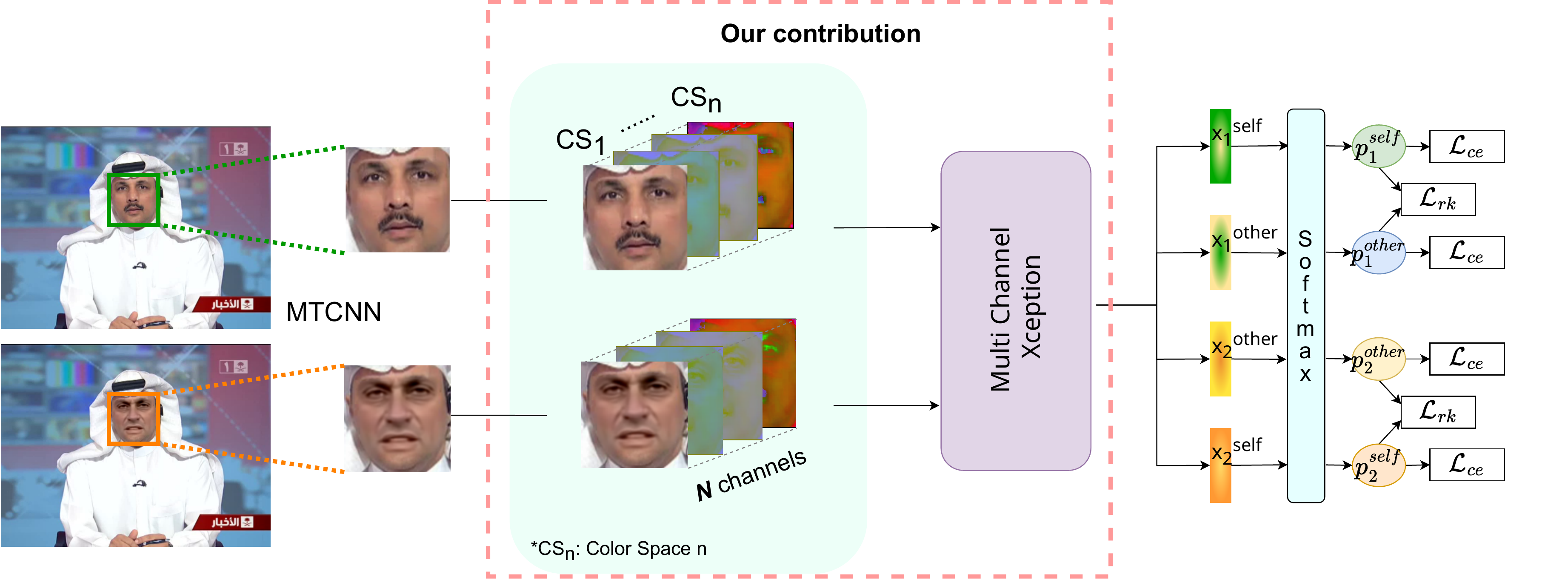}
    \caption{The architecture of MCX-API network.}
    \label{fig:arr_bar}
\end{figure*}

\section{Related Work}
\label{sec:related-works}
\textbf{Deepfakes detection methods.} A track of Deepfakes detection focuses on the unique artifacts on human faces, such as eye blinking~\cite{li2018ictu}, different eye colors~\cite{matern2019exploiting}, abnormal heartbeat rhythms shown on the face~\cite{ciftci2020fakecatcher,fernandes2019predicting}. LipForensics~\cite{haliassos2021lips} targets high-level semantic abnormalities in mouth movements, which the authors observe as a common indicator in many generated videos. Some articles are dedicated to finding inconsistencies in images and videos. These inconsistencies arise out of generation process where landmarks, head pose are inconsistent~\cite{yang2019exposing,agarwal2019protecting} or observable in image blending~\cite{li2018exposing,li2020face}. Cozzolino \textit{et. al.}~\cite{cozzolino2021id} have introduced ID-Reveal, an identity-aware detection approach leveraging a set of reference videos of a target person and trained in an adversarial manner. 
Many papers have utilized CNNs-based methods for detecting features existing in forged images\cite{marra2019gans,rossler2019faceforensics++,nguyen2019use}. Using high-frequency features~\cite{durall2019unmasking,chen2021attentive,qian2020thinking} to distinguish Deepfakes are also gaining more popularity on this topic.
Although pairwise learning have been used for Deepfake detection~\cite{hsu2020deep, xu2022supervised}, they lack the pairwise interactions by using contrastive learning.

\par \textbf{Generalization to unseen manipulations.} While many works are proposed for detecting DeepFakes, they have focused on closed-set experiments where the training and testing set distributions are similar. The open-set experiments indicate that they underperform on unseen manipulations. In the meantime, an increasing number of works have tried to address the problem of generalization of DeepFakes detection. These works have focused on domain adaptation and transfer learning to minimize the task of learning parameters in an end-to-end manner~\cite{aneja2020generalized,kim2021fretal,lee2021tar}. \textit{Cozzolino et. al.}~\cite{cozzolino2018forensictransfer} proposed an autoencoder-like structure ForensicTransfer and the generalization aspect was studied using a single detection method for multiple target domains. The follow-up works like Locality-aware AutoEncoder (LAE)~\cite{du2019towards} and Multi-task Learning were proposed for detecting and segmenting manipulated facial images and videos~\cite{nguyen2019multi}. Transfer learning-based Autoencoder with Residuals (TAR)~\cite{lee2021tar} recently proposed uses the residuals from autoencoders to handle generalizability. \textit{Kim et. al.}~\cite{kim2021fretal} employed the Representation Learning (ReL) and Knowledge Distillation (KD) paradigms to introduce a transfer learning-based Feature Representation Transfer Adaptation Learning (FReTAL) method. While these transfer learning and zero-shot/few-shot learning methods could not wholly deal with the Deepfakes detection generalization problem, because the networks have already seen the manipulated image/videos. Therefore, strictly speaking, it is not an open-set situation.
\par In the meantime, some other novel networks have been proposed dealing with the generalization problem of Deepfakes detection. A new method to detect deepfake images using the cue of the source feature inconsistency within the forged images~\cite{zhao2021learning} is proposed based on the hypothesis that distinct source features can be preserved and extracted through SOTA deepfake generation processes. Joint Audio-Visual Deepfake Detection~\cite{zhou2021joint} is proposed by jointly modeling video and audio modalities. This novel visual/auditory deepfake combined detection task shows that exploiting the intrinsic synchronization between the visual and auditory modalities could benefit deepfake detection. \textit{Xu et. al.}~\cite{xu2022supervised} proposed a novel method using supervised contrastive learning to deal with the generalization problem in detecting forged visual media.

\section{Proposed Method}
\label{sec:proposed-approach}
Fine-grained method has been widely used for classification problems where the categories are visually very similar~\cite{zhuang2020learning, xiao2015application, akata2015evaluation}. We draw similar inspiration to our problem of Deepfake detection following the architecture proposed by earlier~\cite{zhuang2020learning} and build upon with number of improvements. We assert that architecture for fine-grained classification can help in detecting Deepfakes. Unlike the orginal architecture, we introduce Xception~\cite{chollet2017xception} to extract the embeddings motivated by earlier works in Deepfake detection~\cite{rossler2019faceforensics++, zhao2021multi, wang2022m2tr, kim2021fretal}. 

Second, to benefit from information from different color spaces, we make the base network to a multi-channel network. Then, we enforce pairwise learning by following the architecture of Attentive Pairwise Learning \cite{zhuang2020learning}. We propose using the Multi Channel Xception Attentive Pairwise Interaction Network (MCX-API) to deal with the Deepfakes classification problem as detailed further.

\subsection{Architecture}
\par We first utilize MTCNN\cite{zhang2016joint} to crop and align the face region of a single frame. Two selected face images are further sent to a Multi-Channel Xception backbone, and this backbone network extracts two corresponding $\mathrm{D}$-dimension feature vectors $x_{1}$ and $x_{2}$ using the face image represented in $N$ different channels that include RGB, CIELab, HSV, and YCbCr. A mutual vector $x_{m}\in \mathbb{R}^{D}$ is further generated by concatenating $x_{1}$ and $x_{2}$ and using a Multi-Layer Perceptron (MLP) function for mapping $x_{m}$ to get a $\mathrm{D}$ dimension. $x_{m}$ is a joint feature that includes high-level contrastive clues of both input images across multiple color channels.

\par In order to compare $x_{m}$ with $x_{1}$ and $x_{2}$, we need to activate $x_{m}$ using sigmoid function to increase the positive relation with $x_{i}$ and decrease the negative relation against $x_{i}$~\cite{zhuang2020learning}. Therefore, two gate vectors $g_{1}$ and $g_{2}$ will be generated. $g_{i}$ is calculated by $x_{m}$ and $x_{i}$, thus containing contrastive clues and acting as discriminative attention spots semantic contrasts with a distinct view of each $x_{i}$. The gate vector $g_{i}$ is the sigmoid of the output of channel-wise product between $x_{m}$ and $x_{i}$, whose formula is provided in \Cref{eqn:gate-vector}. 
\vspace{-1mm}
\begin{equation}
    g_{i} = sigmoid(x_{m} \odot x_{i}), \;\; i \in{\{1,2\}}
    \label{eqn:gate-vector}
\end{equation}

\par A pairwise interaction between input features $x_{i}$ and gate vectors $g_{i}$ is performed to induce residual attention by comparing one image to the other to distinguish the final class. The sequence of interaction can be shown in \Cref{eqn:pairwise-interaction}.
\begin{equation}
\centering
\begin{split}
    x_{1}^{pristine}=x_{1}+x_{1}\odot g_{1} \\
    x_{1}^{fake}=x_{1}+x_{1}\odot g_{2} \\
    x_{2}^{pristine}=x_{2}+x_{2}\odot g_{2} \\
    x_{2}^{fake}=x_{2}+x_{2}\odot g_{1}
\end{split}
\label{eqn:pairwise-interaction}
\end{equation}
Through the pairwise interaction of each feature $x_{i}$, two attentive feature vectors $x_{i}^{pristine}\in \mathbb{R}^{D}$ and $x_{i}^{fake}\in \mathbb{R}^{D}$ are further produced. The former one is highlighted by its gate vector, and the latter is triggered by the gate vector of the compared image. $x_{i}$ is thus enhanced with discriminative clues from both input features through pairwise interaction.


\subsection{Loss calculation}
The four attentive features $x_{i}^{j}$ where $i \in {\{1,2\}}$ and $j \in {\{pristine,fake\}}$, the pairwise interaction outputs, are fed into a $softmax$ classifier for the loss calculation~\cite{zhuang2020learning}. The output of $softmax$ denoted by $p_{i}^{j}$ is the probability of a feature belonging to a specific class (i.e., non-manipulated or Deepfake). The main loss in our case is the cross-entropy loss 
\begin{equation}
\mathcal{L}_{ce} = -\sum_{i \in \{ 1,2 \}} \sum_{j \in \{ pristine,fake \}} y_{i}^{\intercal} log(p_{i}^{j})
\label{eqn:lce}
\end{equation}
where $y_{i}$ is the one-hot label for image $i$ in the pair and $\intercal$ represents the transpose. MCX-API can be trained to determine all the attentive features $x_{i}^{j}$ under the supervision of the label $y_{i}$ through this loss.

\par Furthermore, a hinge loss of score ranking regularization
\begin{equation}
    \mathcal{L}_{rk} = \sum_{i\in {\{1, 2\}}} max(0, p_{i}^{fake}(c_{i})-p_{i}^{pristine}(c_{i})+\epsilon )
\label{eqn:lrk}
\end{equation}
is also introduced when computing the complete loss~\cite{zhuang2020learning}. $c_{i}$ is the corresponding index associated with the ground truth label of image $i$. So $p_{i}^{j}(c_{i})$ is a softmax score of $p_{i}^{j}$. Since $x_{i}^{pristine}$ is activated by its gate vector $g_{i}$, it should contain more discriminative features to identify the corresponding image, compared to $x_{i}^{fake}$. $\mathcal{L}_{rk}$ is utilized to promote the priority of $x_{i}^{pristine}$ where the score difference between $p_{i}^{fake}(c_{i})$ and $p_{i}^{pristine}(c_{i})$ should be greater than a margin.
The whole loss for a pair is composed of two losses, cross-entropy loss $\mathcal{L}_{ce}$ and score ranking regularization $\mathcal{L}_{rk}$ with coefficient $\lambda$. 
\begin{equation}
    \mathcal{L} = \mathcal{L}_{ce} + \lambda \mathcal{L}_{rk}
\label{eqn:loss}
\end{equation}
In this way, MCX-API is able to take feature priorities into account adaptively and learns to recognize each image in the pair.

\section{Experiments and Results}
\label{sec:Resutls}

\subsection{Datasets}
\textbf{Training data: }We select FaceForensics++ \cite{rossler2019faceforensics++} to train the proposed approach. This forensics dataset consists of 1000 original videos and corresponding number of manipulated videos consisting of 1000 videos for each of the subsets - DeepFakes (denoted as DF)~\cite{ffdf}, Face2Face (denoted as F2F)~\cite{thies2016face2face}, FaceSwap (denoted as FS)~\cite{fffs}, and NeuralTextures (denoted as NT)~\cite{thies2019deferred}.

\textbf{Cross-dataset Validation: }We also select three other SOTA datasets for generalization test and comparison. \textbf{Celeb-DF~\cite{li2020celeb}}: For Celeb-DF, we choose id51-id61 from Celeb-real, Celeb-synthesis and id240-id299 from YouTube-real for the test set.
\textbf{KoDF~\cite{kwon2021kodf}} We randomly selected 265 real videos and 734 fake ones as our test set.
\textbf{FakeAV~\cite{khalid2021fakeavceleb}} We randomly selected 500 videos as our test set.


\begin{table*}[htp]
\caption{\textbf{Frame-level BOSC Accuracy and AUC for our proposed MCX-API networks and SOTA methods on seen data.} We compare the results with the SOTA methods on DF/F2F/FS/NT respectively. All networks are trained on the whole FF++ c23 dataset. The data of the first three methods are adopted from Table 5 in Appendix of FF++~\cite{chollet2017xception}.}
\label{tab:bosc-ffpp}
\begin{threeparttable}
\centering
\begin{tabular}{llllllll}
\toprule
FF++ c23 &  & \multicolumn{6}{c}{Frame-level (BOSC(\%)/AUC)} \\ \cline{1-1} \cline{3-8} 
Method &  & \multicolumn{1}{c}{DF} & \multicolumn{1}{c}{F2F} & \multicolumn{1}{c}{FS} & \multicolumn{1}{c}{NT} &  & Average \\ \cline{1-6} \cline{8-8} 
Cozzolino \textit{et al.}~\cite{cozzolino2017recasting} &  & 75.51/ - & 86.34/ - & 76.81/ - & 75.34/ - &  & 78.50/ - \\
Bayar and Stamm~\cite{bayar2016deep} &  & 90.25/ - & 93.96/ - & 87.74/ - & 83.69/ - &  & 88.91/ - \\
MesoNet~\cite{afchar2018mesonet} &  & 89.55/ - & 88.60/ - & 81.24/ - & 92.19/ - &  & 87.90/ - \\
Xception\tnote{*} \cite{chollet2017xception} &  & 96.35/0.9941 & 96.26/0.9937 & 96.29/0.9952 & 92.43/0.9736 &  & 95.33/0.9892 \\
SupCon\tnote{*} \cite{xu2022supervised} &  & 97.18/0.9984 & 96.88/0.9978 & 97.05/0.9980 & 92.92/0.9846 &  & 96.01/0.9947 \\ 
API-Net(ResNet101)\tnote{*} \cite{zhuang2020learning} &  & 88.71/0.9820 & 90.13/0.9860 & 87.79/0.9728 & 82.96/0.9248 &  & 87.40/0.9664 \\ \hline
Ours &  &  &  &  &  &  &  \\
\textbf{MCX-API(RGB)} &  & \textbf{98.75}/0.9996 & \textbf{99.90}/0.9986 & \textbf{98.5}/\textbf{0.9993} & \textbf{96.75}/0.9896 &  & \textbf{98.48}/0.9968 \\
\textbf{MCX-API(RGB+HSV)} &  & \textbf{98.75}/0.9988 & 98.50/0.9979 & 97.75/0.9978 & 95.75/0.9829 &  & 97.69/0.9943 \\
\textbf{MCX-API(RGB+CIELab)} &  & 97.00/0.9996 & 96.50/0.9985 & 96.25/0.9989 & 95.25/0.9909 &  & 96.25/0.9970 \\
\textbf{MCX-API(RGB+YCbCr)} &  & 98.00/\textbf{0.9998} & 98.25/\textbf{0.9991} & 97.75/\textbf{0.9993} & \textbf{96.75}/0.9920 &  & 97.69/\textbf{0.9976} \\
\textbf{MCX-API(RGB+HSV+CIELab)} &  & 96.50/0.9990 & 95.50/0.9888 & 96.00/0.9835 & 95.50/\textbf{0.9933} &  & 95.88/0.9912 \\
\textbf{MCX-API(RGB+LAB+YCbCr)} &  & 92.00/0.9963 & 92.25/0.9972 & 91.50/0.9960 & 91.00/0.9870 &  & 91.69/0.9941 \\
\bottomrule
\end{tabular}
    \begin{tablenotes}
         \footnotesize
         \item [*] Our implementation of the method.
    \end{tablenotes}
\end{threeparttable}
\end{table*}

\textbf{Implementation details.} 
We choose uncompressed videos for our experiments in this work using the Pytorch framework~\cite{pytorch} to develop the models and the experiments are conducted on Python 3.6 environment on NVIDIA Tesla V100 32Gb in IDUN system owned by NTNU~\cite{sjalander+:2019epic}. 

Multi-task Cascade Convolutional Neural Networks (MTCNN)~\cite{zhang2016joint} is employed for face detection and face alignment since our experiments are focused on detecting the manipulated face region alone. We allow loose cropping of the face region to capture the entire silhouette against tight cropping. The first 30 frames from each video are extracted, resulting in 150000 total images. We use random cropping in the training phase and center cropping during the testing phase ($512^{2}\to448^{2}$). In all our experiments, we employ Xception as the backbone where we derive the feature vector $x_{i}\in \mathbb{R}^{2048}$ after the global average pooling. We use a batch sampler during the training by randomly sampling three categories in each batch. For each category, we randomly choose nine images due to the limitations of the GPU and memory constraints. We further exercise care to have no sample overlap among all batches, as we exclude the selected sample from the dataset. We locate its most similar image from both its own class and the rest classes for each image by calculating the distance between features by utilizing both Euclidean distance and cosine distance. Each image would get one image as its intra- and inter-pair in the batch, respectively. Each pair is used as input $x_{1}$ and $x_{2}$ as well as generating a mutual vector $x_{m}\in \mathbb{R}^{2048}$ through the concatenation and the multilayer perceptron (MLP).
\par Based on empirical evaluations, we adopt the coefficient $\lambda$ in \Cref{eqn:loss} as 1.0, and 0.05 as the margin value in the score-ranking regularization. We use cosine annealing strategy to alter the learning rate starting from 0.01 \cite{zhao2021learning}. We train the network with 100 epochs and freeze the parameters in the CNN backbone, and further on train only the classifier in the first eight epochs.

\textbf{Evaluation Metrics.}
We adopt Balanced-Open-Set-Classification (BOSC) accuracy and AUC as evaluation metrics.
$BOSC = \frac{Sensitivity+Specificity}{2}$, where $Sensitivity=\frac{TP}{TP + FN}$ and $Specificity = \frac{TN}{TN + FP}$.

\begin{table}[htp]
\caption{Comparison of the test results on the FF++ dataset with c23 (high-quality compression) settings. Training for all networks is carried out on FF++ c23. The accuracy and AUC score are at frame-level. The best performances are marked in bold. Data for Xception, $F^3$-Net, and EfficientNet-B4 are adopted from Table 2 in MaDD~\cite{zhao2021multi}.}
\label{tab:ff-sota}
\centering
\begin{tabular}{llcc}
\toprule
Method          && ACC            & AUC            \\ \cline{1-1} \cline{3-4} 
Xception        && 95.73          & 0.9909          \\ 
$F^3$-Net~\cite{qian2020thinking}          && 97.52          & 0.9810          \\ 
EfficientNet-B4~\cite{tan2019efficientnet} && 96.63          & 0.9918          \\ 
DCL~\cite{sun2022dual}             && 96.74          & 0.9930          \\ 
MaDD~\cite{zhao2021multi}            && 97.60          & 0.9929          \\ 
M2TR~\cite{wang2022m2tr}            && 97.93          & 0.9951          \\ 
API-Net            && 87.40          & 0.9664          \\ \hline
Ours            && \textbf{98.48} & \textbf{0.9968} \\
\bottomrule
\end{tabular}
\end{table}

\begin{table*}[htp]
\caption{\textbf{Video-level BOSC Accuracy and AUC for our proposed MCX-API networks and SOTA methods on unseen data.} We compare the results with the SOTA methods on FakeAV/KoDF/Celeb-DF respectively. All the networks are trained on the whole FF++ c23 dataset. The data of the SOTA methods are adopted from Table 2 from \cite{cozzolino2022audio}.}
\label{tab:cross-all}
\centering
\begin{threeparttable}
\begin{tabular}{lllll}
\toprule
FF++ c23 &  & \multicolumn{3}{c}{Video-level (BOSC(\%)/AUC)} \\ \cline{1-1} \cline{3-5} 
Method &  & \multicolumn{1}{c}{FakeAV} & \multicolumn{1}{c}{KoDF} & \multicolumn{1}{c}{Celeb-DF} \\ \hline
Xception\tnote{*} &  & 23.99/0.450 & 25.97/0.482 & 31.34/0.505 \\
Seferbekov~\cite{seferbekov} &  & \textbf{95.0/0.986} & 79.2/0.884 & 75.3/0.860 \\
FTCN~\cite{zheng2021exploring} &  & 64.9/0.840 & 63.0/0.765 & - \\
LipForensics~\cite{haliassos2021lips} &  & 83.3/0.976 & 56.1/0.929 & -/0.820 \\
ID-reveal~\cite{cozzolino2021id} &  & 63.7/0.876 & 60.3/0.702 & 71.6/0.840 \\
POI~\cite{cozzolino2022audio} &  & 86.6/0.941 & 81.1/0.899 & - \\ 
API-Net(ResNet101)\tnote{*} &  & 59.99/0.72 & 66.92/0.76 & 58.00/0.76 \\\hline
Ours &  &  &  &  \\
\textbf{MCX-API(RGB)} &  & 74.94/0.95    & 78.09/\underline{0.87}  & 77.88/0.87 \\
\textbf{MCX-API(HSV)} &  & 74.63/0.75 & \underline{80.64}/0.85  & 75.67/0.88 \\
\textbf{MCX-API(CIELab)} &  & 84.28/0.90 & \textbf{81.16/0.90}  & 64.28/0.81 \\
\textbf{MCX-API(RGB+HSV)} &  & 71.58/0.93    & 78.11/\underline{0.87}  & \underline{80.18}/0.88 \\
\textbf{MCX-API(RGB+CIELab)} &  & 83.89/0.93 & 77.93/0.83  & 68.34/\textbf{0.91} \\
\textbf{MCX-API(RGB+YCbCr)} &  & 70.41/0.92   & 78.39/0.85  & \textbf{90.87}/\underline{0.90} \\
\textbf{MCX-API(RGB+HSV+CIELab)} &  & \underline{92.38/0.98} & 78.91/0.83 & 59.04/0.89 \\
\textbf{MCX-API(RGB+LAB+YCbCr)} &  & 82.93/0.96  & 76.20/0.80  &  54.92/0.85 \\
\bottomrule
\end{tabular}
    \begin{tablenotes}
         \footnotesize
         \item [*] Our implementation of the method.
    \end{tablenotes}
\end{threeparttable}
\end{table*}

\subsection{Experimental Results}
\label{ExperimentalResults}
We evaluate the effectiveness of the proposed MCX-API network with both seen and unseen data in this section.  

\subsubsection{Intra-dataset Evaluation (Closed Set Protocol)}
We conduct experiments on six networks with different color spaces on MCX-API whose results are presented in \cref{tab:bosc-ffpp}. All networks are trained with all four manipulated methods along with pristine in FF++ c23 dataset. We test the frame-level detection performance on the test data of FF++ c23 in a non-overlapping manner regarding the ID. 

In \cref{tab:bosc-ffpp}, the frame-level test results are listed. We observe that our proposed MCX-API network with RGB inputs reaches the highest average accuracy, 98.48\%. In addition, this setting also gains the highest accuracy on DF, F2F, and FS with 98.87\%, 99.90\% and 98.50\%, respectively. MCX-API with YCbCr achieves the highest accuracy for NT with 97.00\%. As RGB provides best performance under  3-channel setting, we combine RGB with HSV, CIELab, and YCbCr, respectively, to create three 6-channel MCX-API networks. From the second block in \cref{tab:bosc-ffpp}, we can see that RGB+YCbCr obtains the highest average AUC score of 0.9976 and the best performance on DF, F2F, and FS regarding AUC score. This indicates better prediction output scores using MCX-API with the combination of RBG and YCbCr color spaces. The 9-channel MCX-API network with RGB, HSV, and CIELab further gains the highest 0.9933 AUC score for NT.

The results of the comparison with the SOTA methods are reported in \cref{tab:ff-sota}. All networks are trained on FF++ c23 (high-quality compression). The accuracy and AUC scores are measured at frame level. The results are averaged on all the test sets from FF++ c23, including pristine and all four kinds of manipulated videos. Our proposed method MCX-API with RGB color space obtains the best performance compared to SOTA methods. The best accuracy of the BOSC is 98.48\%, and the highest AUC score is 0.9968. The result shows that our idea of pairwise learning in a fine-grained manner could work well in inter-class (closed-set) setting of Deepfake detection problem.

\subsubsection{Cross-dataset Evaluation}
We conduct a comparison on cross-dataset validation with SOTA methods to validate the proposed approach.  We employ FakeAV, KoDF, and Celeb-DF to test the generalizability of our MCX-API network. Training for all networks are carried out on the FF++ c23 dataset and tested on FakeAV, KoDF, and Celeb-DF. We note that MCX-API with CIELab color space gets the best scores for KoDF with an accuracy of 81.86\% and an AUC score of 0.90 as presented in \cref{tab:cross-all}. MCX-API with RGB+YCbCr wins in the cross-dataset validation for Celeb-DF with an accuracy of 90.87\% and the second best AUC score 0.90. MCX-API with color space RGB+HSV+CIELab achieves the second best place for FakeAV with 92.38\% accuracy and 0.98 AUC score. In general, our proposed network gets a relatively better performance than the SOTA methods which indicates the better generalizability of the proposed MCX-API network.

\section{Explainable Analysis of MCX-API}
\label{sec:ExplainableAnalysisofMCX-API}
We further analyze the network to understand the performance gain by analyzing embeddings using t-SNE plots~\cite{van2008visualizing} and class activation maps~\cite{selvaraju2017grad, chattopadhay2018grad, draelos2020use, jiang2021layercam, fu2020axiom}. While the t-SNE provides topology explanations of the learned features, the activation maps allow for a better visualization of what has been learned by our network. 

\begin{figure*}[ht]\centering
	\resizebox{2.\columnwidth}{!}{
		\begin{tabular}{cc}
			\begin{tikzpicture}[spy using outlines={rectangle,yellow,magnification=3,size=4.0cm, connect spies, every spy on node/.append style={very thick}}]
				\node {\includegraphics[width=9cm, height=9cm]{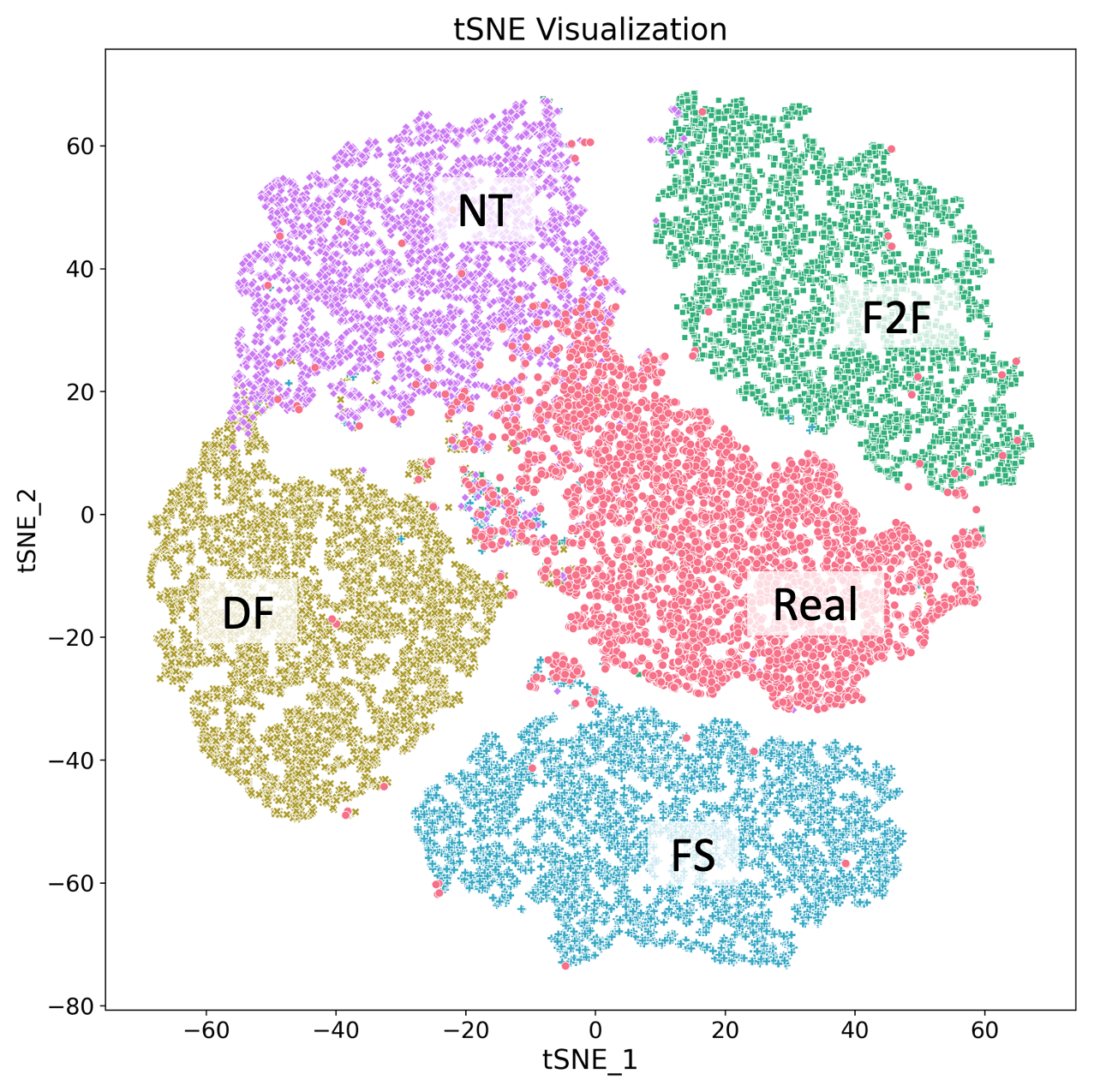}};
				\spy[red] on (-0.3,0.2) in node [right] at (-5.5,6.3);
				\spy[green,size=2.7cm] on (0.1,-1.2) in node [right] at (-6.9,-2.2);
				\spy[blue,size=3.cm] on (0.5,2.1) in node [right] at (1.,6.);
                    \spy[orange,size=2.7cm] on (-2.5,1.0) in node [left] at (-4.2,1);
			\end{tikzpicture} & 

			\begin{tikzpicture}[spy using outlines={rectangle,yellow,magnification=3,size=4cm, connect spies, every spy on node/.append style={very thick}}]
				\node {\includegraphics[width=9cm, height=9cm]{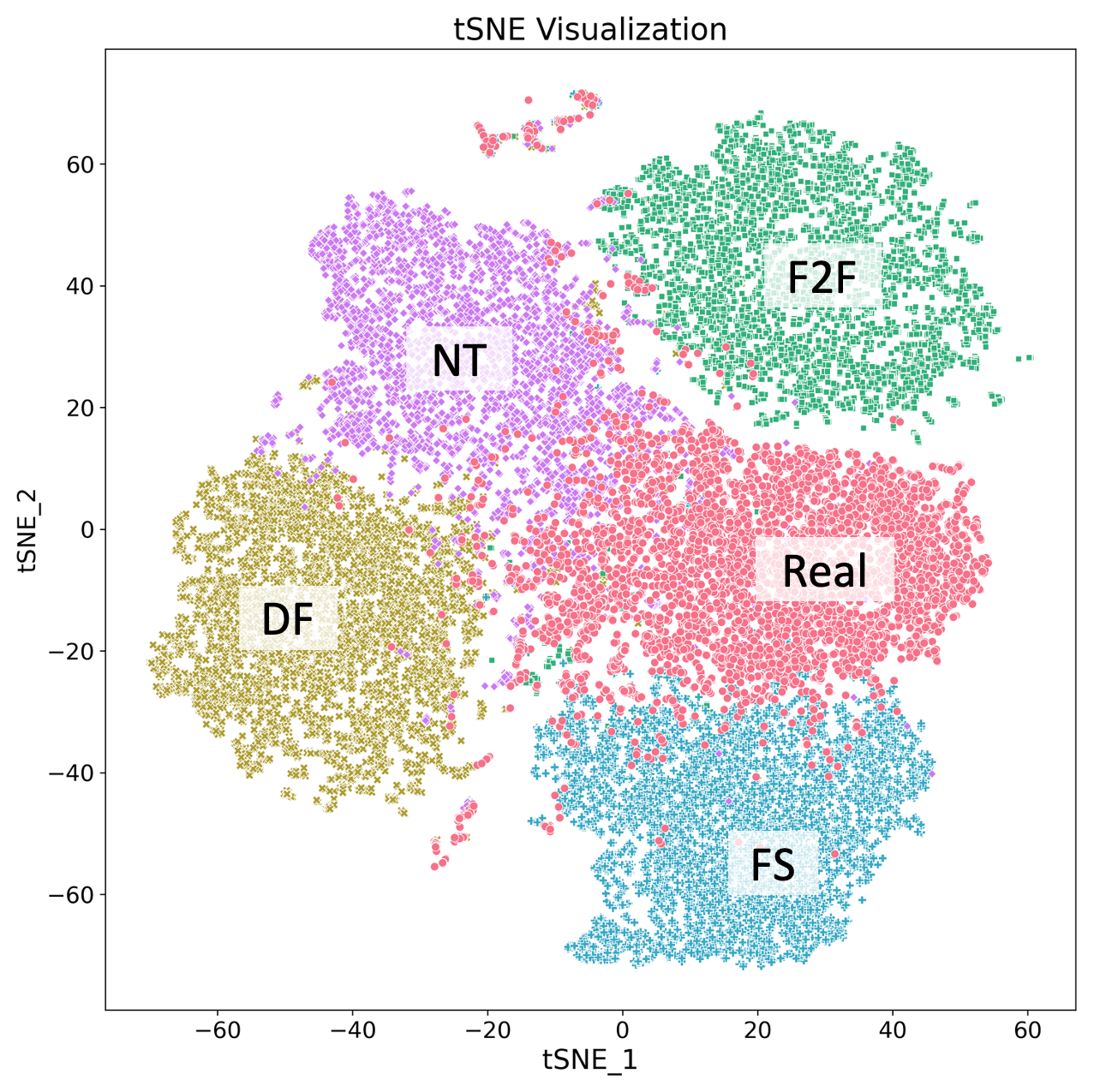}};
				\spy[red] on (-0.3,0.2) in node [right] at (-5.5,6.3);
				\spy[green,size=2.7cm] on (0.1,-1.2) in node [right] at (-6.9,-2.2);
				\spy[blue,size=3.cm] on (0.5,2.1) in node [right] at (1.,6.);
                    \spy[orange,size=2.7cm] on (-2.,1.0) in node [left] at (-4.2,1);
			\end{tikzpicture}\\
   
			\large{MCX-API (RGB)} & \large{API} 
		\end{tabular}
	}
	\caption{Data visualizations in 2D by t-SNE for MCX-API(RGB) and API. The left plot is t-SNE for our proposed MCX-API. The right plot is t-SNE for base architecture API-Net. We blow up the intersection parts and outliers for a clear view.}
\label{fig:tsne}
\end{figure*}

\begin{figure*}[ht]\centering
	\resizebox{1.7\columnwidth}{!}{
		\begin{tabular}{ccc}
			\begin{tikzpicture}[spy using outlines={circle,yellow,magnification=3,size=4.5cm, connect spies, every spy on node/.append style={ultra thick}}]
				\node {\includegraphics[width=6cm, height=6cm]{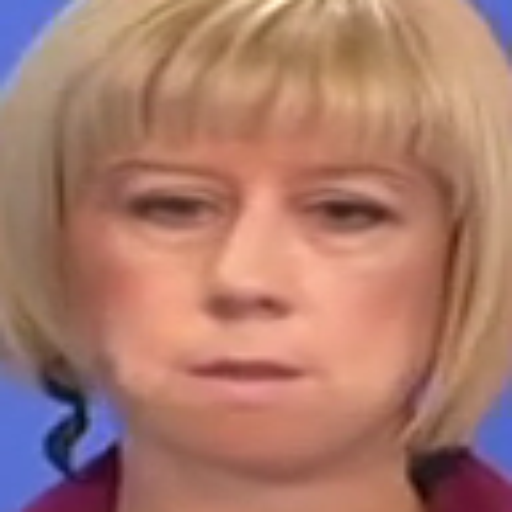}};
				\spy[red] on (-0.1,-0.4) in node [right] at (2.,-4);
				\spy[yellow] on (-1.4,-1.3) in node [right] at (-5,-5);
				\spy[blue] on (1.6,0.8) in node [right] at (2,4);
			\end{tikzpicture} & 
			
			\begin{tikzpicture}[spy using outlines={circle,yellow,magnification=3,size=4.5cm, connect spies, every spy on node/.append style={ultra thick}}]
				\node {\includegraphics[width=6cm, height=6cm]{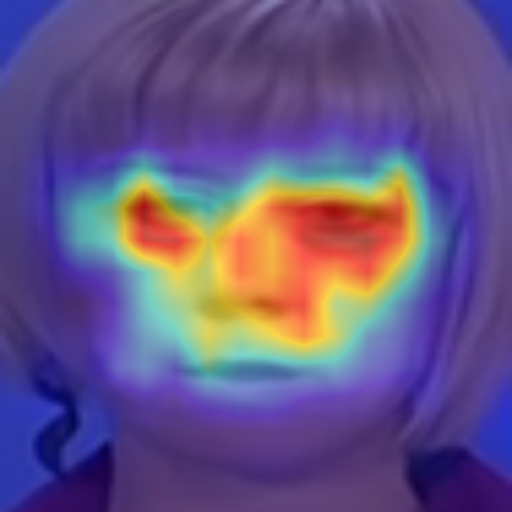}};
				\spy[red] on (-0.1,-0.4) in node [right] at (2.,-4);
				\spy[yellow] on (-1.4,-1.3) in node [right] at (-5,-5);
				\spy[blue] on (1.6,0.8) in node [right] at (2,4);
			\end{tikzpicture} & 
   
			\begin{tikzpicture}[spy using outlines={circle,yellow,magnification=3,size=4.5cm, connect spies, every spy on node/.append style={ultra thick}}]
				\node {\includegraphics[width=6cm, height=6cm]{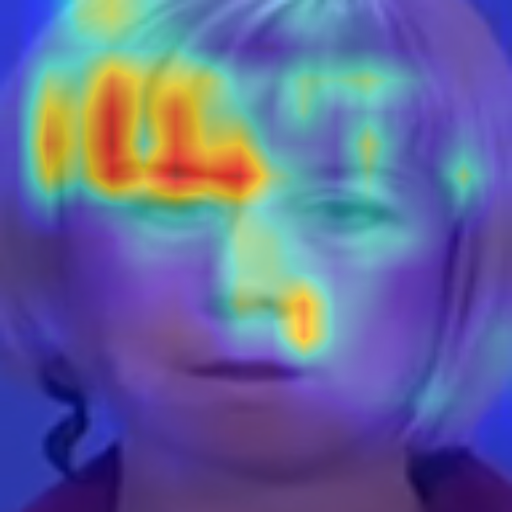}};
				\spy[red] on (-0.1,-0.4) in node [right] at (2.,-4);
				\spy[yellow] on (-1.4,-1.3) in node [right] at (-5,-5);
				\spy[blue] on (1.6,0.8) in node [right] at (2,4);
			\end{tikzpicture}\\
   
			\huge{DF} &\huge{MCX-API (RGB)} & \huge{API} \\

   			\begin{tikzpicture}[spy using outlines={circle,yellow,magnification=3,size=4.5cm, connect spies, every spy on node/.append style={ultra thick}}]
				\node {\includegraphics[width=6cm, height=6cm]{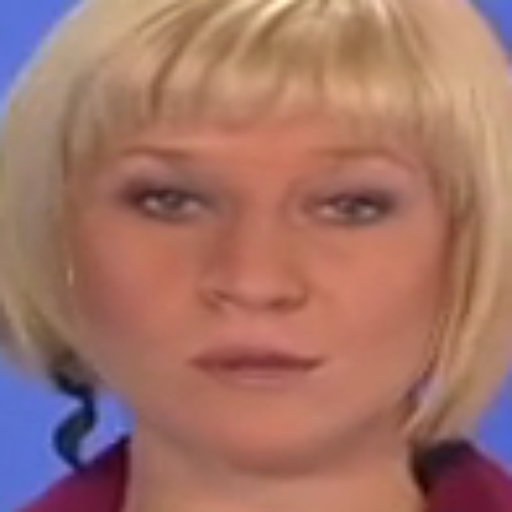}};
				\spy[red] on (-1.,1.) in node [left] at (-2.7,2.);
				\spy[yellow] on (0.,-1.3) in node [right] at (-5,-5);
				\spy[blue] on (1.6,0.8) in node [right] at (2,4);
			\end{tikzpicture} & 
			
			\begin{tikzpicture}[spy using outlines={circle,yellow,magnification=3,size=4.5cm, connect spies, every spy on node/.append style={ultra thick}}]
				\node {\includegraphics[width=6cm, height=6cm]{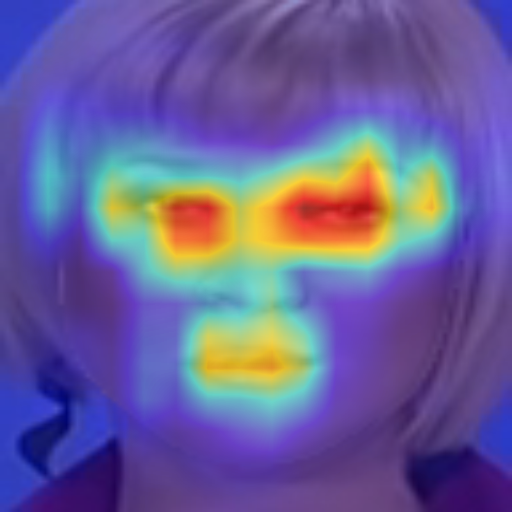}};
				\spy[red] on (-1.,1.) in node [left] at (-2.7,2.);
				\spy[yellow] on (0.,-1.3) in node [right] at (-5,-5);
				\spy[blue] on (1.6,0.8) in node [right] at (2,4);
			\end{tikzpicture} & 
   
			\begin{tikzpicture}[spy using outlines={circle,yellow,magnification=3,size=4.5cm, connect spies, every spy on node/.append style={ultra thick}}]
				\node {\includegraphics[width=6cm, height=6cm]{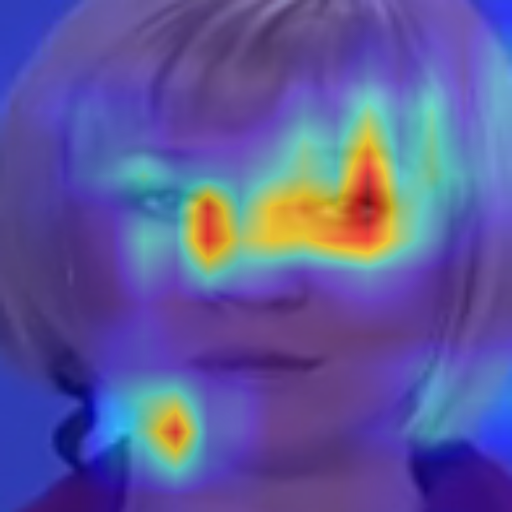}};
				\spy[red] on (-1.,1.) in node [left] at (-2.7,2.);
				\spy[yellow] on (0.,-1.3) in node [right] at (-5,-5);
				\spy[blue] on (1.6,0.8) in node [right] at (2,4);
			\end{tikzpicture}\\
   
			\huge{F2F} &\huge{MCX-API (RGB)} & \huge{API} 
   
		\end{tabular}
	}
	\caption{Blow up in activation maps from LayerCAM analysis of MCX-API(RGB) and base architecture API-Net on DF and F2F faces.}
	\label{fig:blowup-activation}
\end{figure*}

\begin{figure}[htp]
    \centering
    \subfigure[Visualization of the last block of the exit flow of MCX-API (RGB).]{\label{fig:visualization-MCXapi}\includegraphics[width=0.95\linewidth]{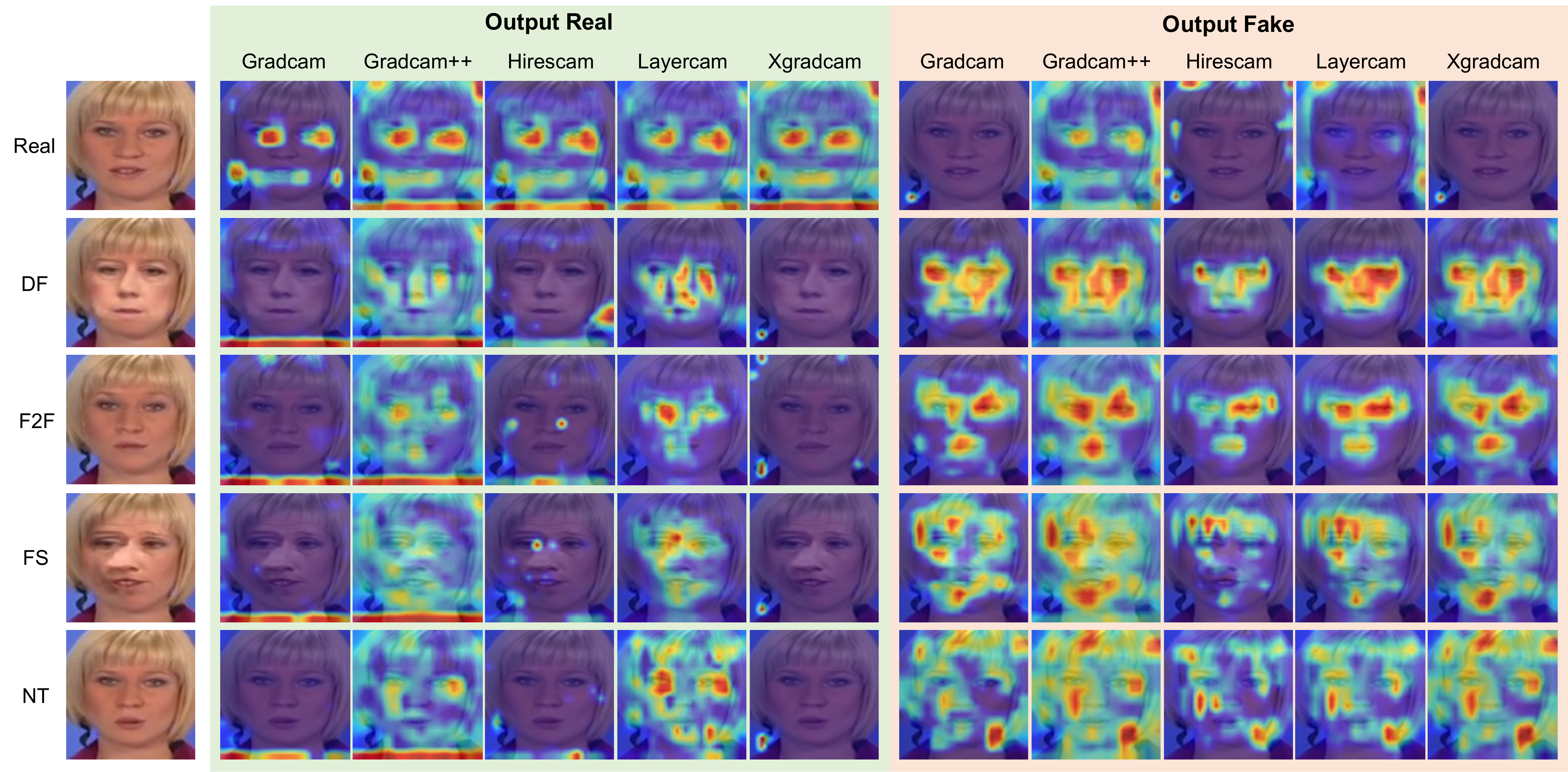}}
    \subfigure[Visualization of the last block of the API-Net.]{\label{fig:visualization-api}\includegraphics[width=0.95\linewidth]{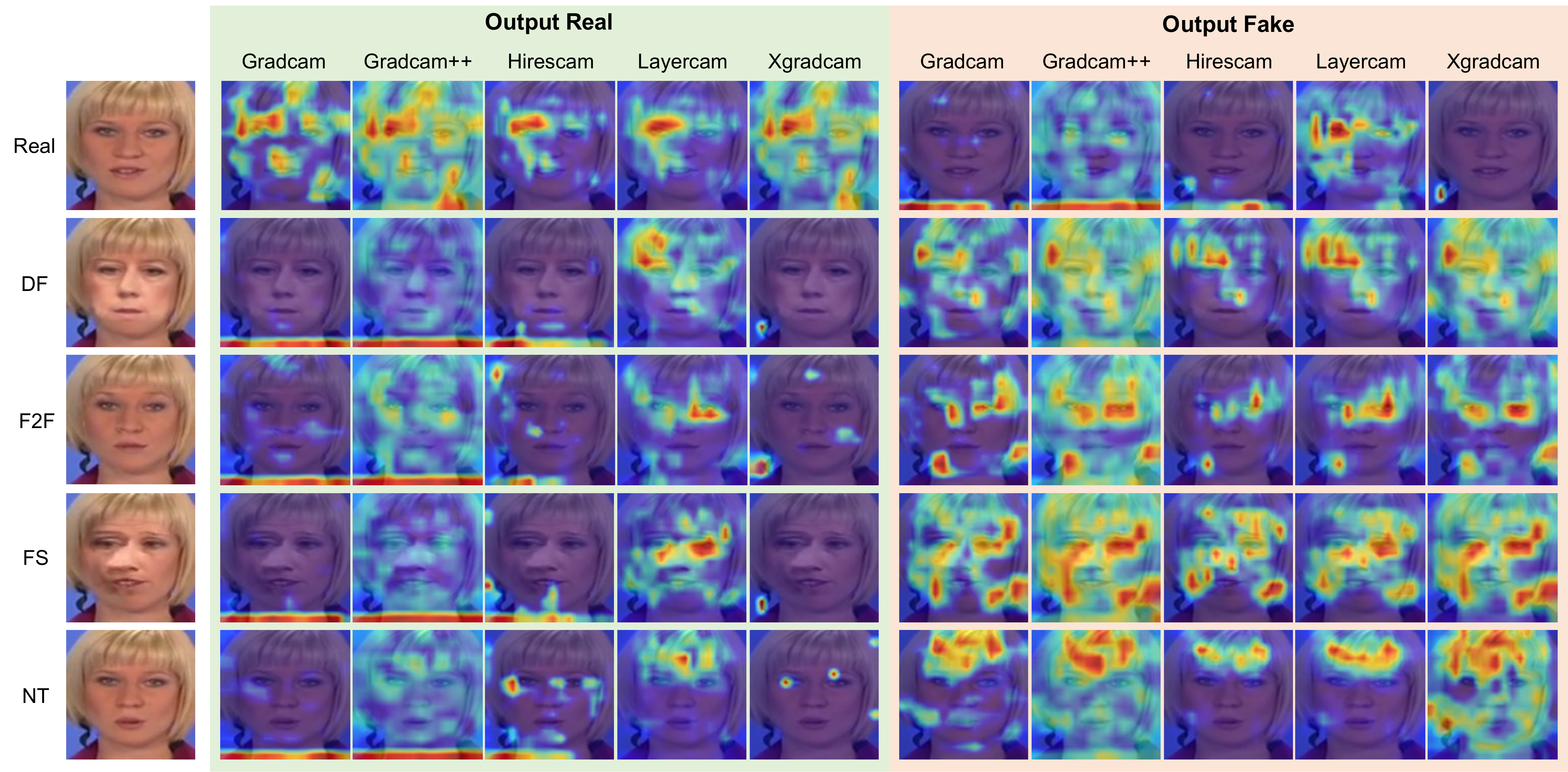}}
    
    \caption{Visualization of the last layer of MCX-API (RGB) and API networks. We utilize Grad-CAM~\cite{selvaraju2017grad}, Grad-CAM++~\cite{chattopadhay2018grad}, HiResCAM~\cite{draelos2020use}, LayberCAM~\cite{jiang2021layercam} and XGradCAM~\cite{fu2020axiom} as our visualization tool. For larger figure, please refer to \cref{fig:activation-map-large}.}
    \label{fig:activation-map}
\end{figure}

\subsection{Data Visualizations With t-SNE}
The results of a t-SNE 2D map for the feature vectors are illustrated in \cref{fig:tsne}. We compare the t-SNE of our MCX-API and base architecture API-Net.
We notice that the five classes of Real, DF, F2F, FS, and NT for MCX-API are well separated with five different clusters as against the base architecture of API-Net. There is an unclear boundary between Real and NT, shown in the blue box for MCX-API. This overlapping can be the reason for the relatively lower accuracy obtained on NT. There are small areas overlapping between DF/NT(yellow/purple) and Real/F2F(red/blue). We further notice a few samples of Real (red dots) distributed in each fake class, leading to the errors of our proposed network.

\subsection{Visualizing Decisions With Attention Maps}
We apply different class activation visualization methods on the last layer of proposed network to analyze MCX-API shown in \cref{fig:activation-map}. For comparison, we also show the visualization of the base API-Net. Precisely, we adopt Grad-CAM~\cite{selvaraju2017grad}, Grad-CAM++~\cite{chattopadhay2018grad}, HiResCAM~\cite{draelos2020use}, LayberCAM~\cite{jiang2021layercam} and XGradCAM~\cite{fu2020axiom}. The visualization results are provided in \cref{fig:visualization-MCXapi} for our proposed MCX-API and in \cref{fig:visualization-api} for API-Net.

The activation map for Output Real is on the left part with a green background, and the activation map for Output Fake is on the right part with a pink background. The rows from top to bottom are the visualization for five classes of Real, DF, F2F, FS, and NT, respectively. We can observe that real images gains more attention within Output Real(left part) than Output Fake(right part). In contrast, fake images obtain more attention within Output Fake than Output Real. This explains the ability of our network to detect Deepfakes.

We further blow up the activation maps from LayerCAM for DF and F2F images in \cref{fig:blowup-activation}. From visual analysis, it is evident that the MCX-API focuses more on the facial region, such as the eyes and the mouth. For instance, double eyebrows are found in the DF image (blue circle). MCX-API pays more attention than API around this region. 

\section{Limitations of our work}
We notice in \cref{tab:bosc-ffpp} that with the increase in color spaces, there are no apparent improvements in BOSC accuracy. We assume that there is redundant information among channels, and further work would be focused on finding the most helpful color information to extend our proposed approach. We also observe that no single configuration could perform reasonably well for all the unseen data, which is the biggest issue for Deepfake detection field. Introducing other information, such as temporal data and audio, would be a good idea as more inconsistency could be found by extending our work to video based approach.

\section{Conclusion}
\label{sec:conclusion}
There is an imperative need for a generalized DeepFakes detection method to deal with the newer manipulation methods in visual media. In this paper, we proposed to apply the Multi-Channel Xception Attentive Pairwise Interaction (MCX-API) network to the Deepfakes detection field in a fine-grained manner. We conducted experiments on the publicly available FaceForensics++ dataset, and our approach obtained better performance than the SOTA approaches on both seen and unseen manipulation types. We obtain 98.48\% BOSC accuracy on the FF++ dataset and 90.87\% BOSC accuracy on the CelebDF dataset suggesting a promising direction for the generalization of DeepFake detection. Comprehensive ablation studies have been conducted to understand our algorithm better. We further explain the performance of our network by using t-SNE and attention maps. The results showed that Deepfake had been well separated from real videos. While our approach has indicated a promising solution to obtain a generalized detection mechanism, we have listed certain limitations that can pave the way for future work.  

{\small
\bibliographystyle{ieee_fullname}
\bibliography{egbib}

\begin{thebibliography}{10}\itemsep=-1pt

\bibitem{DeepFaceLab}
Deepfacelab.
\newblock \url{https://github.com/iperov/DeepFaceLab}.

\bibitem{ffdf}
Deepfakes-faceswap.
\newblock \url{https://github.com/deepfakes/faceswap}.
\newblock 2021-10-25.

\bibitem{FaceApp}
Faceapp - most popular selfie editor.
\newblock \url{https://www.faceapp.com/}.

\bibitem{fffs}
Marekkowalski-faceswap.
\newblock \url{https://github.com/MarekKowalski/FaceSwap}.
\newblock 2021-10-25.

\bibitem{pytorch}
Pytorch.
\newblock \url{https://pytorch.org/}.

\bibitem{seferbekov}
Seferbekov, s.: Deepfake detection (dfdc) team sefer.
\newblock \url{https://github.com/selimsef/dfdc_deepfake_challenge/}.

\bibitem{afchar2018mesonet}
Darius Afchar, Vincent Nozick, Junichi Yamagishi, and Isao Echizen.
\newblock Mesonet: a compact facial video forgery detection network.
\newblock In {\em 2018 IEEE international workshop on information forensics and
  security (WIFS)}, pages 1--7. IEEE, 2018.

\bibitem{agarwal2019protecting}
Shruti Agarwal, Hany Farid, Yuming Gu, Mingming He, Koki Nagano, and Hao Li.
\newblock Protecting world leaders against deep fakes.
\newblock In {\em CVPR workshops}, volume~1, 2019.

\bibitem{akata2015evaluation}
Zeynep Akata, Scott Reed, Daniel Walter, Honglak Lee, and Bernt Schiele.
\newblock Evaluation of output embeddings for fine-grained image
  classification.
\newblock In {\em Proceedings of the IEEE conference on computer vision and
  pattern recognition}, pages 2927--2936, 2015.

\bibitem{aneja2020generalized}
Shivangi Aneja and Matthias Nie{\ss}ner.
\newblock Generalized zero and few-shot transfer for facial forgery detection.
\newblock {\em arXiv preprint arXiv:2006.11863}, 2020.

\bibitem{bayar2016deep}
Belhassen Bayar and Matthew~C Stamm.
\newblock A deep learning approach to universal image manipulation detection
  using a new convolutional layer.
\newblock In {\em Proceedings of the 4th ACM workshop on information hiding and
  multimedia security}, pages 5--10, 2016.

\bibitem{chattopadhay2018grad}
Aditya Chattopadhay, Anirban Sarkar, Prantik Howlader, and Vineeth~N
  Balasubramanian.
\newblock Grad-cam++: Generalized gradient-based visual explanations for deep
  convolutional networks.
\newblock In {\em 2018 IEEE winter conference on applications of computer
  vision (WACV)}, pages 839--847. IEEE, 2018.

\bibitem{chen2021attentive}
Zehao Chen and Hua Yang.
\newblock Attentive semantic exploring for manipulated face detection.
\newblock In {\em ICASSP 2021-2021 IEEE International Conference on Acoustics,
  Speech and Signal Processing (ICASSP)}, pages 1985--1989. IEEE, 2021.

\bibitem{chollet2017xception}
Fran{\c{c}}ois Chollet.
\newblock Xception: Deep learning with depthwise separable convolutions.
\newblock In {\em Proceedings of the IEEE conference on computer vision and
  pattern recognition}, pages 1251--1258, 2017.

\bibitem{ciftci2020fakecatcher}
Umur~Aybars Ciftci, Ilke Demir, and Lijun Yin.
\newblock Fakecatcher: Detection of synthetic portrait videos using biological
  signals.
\newblock {\em IEEE Transactions on Pattern Analysis and Machine Intelligence},
  2020.

\bibitem{cozzolino2022audio}
Davide Cozzolino, Matthias Nie{\ss}ner, and Luisa Verdoliva.
\newblock Audio-visual person-of-interest deepfake detection.
\newblock {\em arXiv preprint arXiv:2204.03083}, 2022.

\bibitem{cozzolino2017recasting}
Davide Cozzolino, Giovanni Poggi, and Luisa Verdoliva.
\newblock Recasting residual-based local descriptors as convolutional neural
  networks: an application to image forgery detection.
\newblock In {\em Proceedings of the 5th ACM workshop on information hiding and
  multimedia security}, pages 159--164, 2017.

\bibitem{cozzolino2021id}
Davide Cozzolino, Andreas R{\"o}ssler, Justus Thies, Matthias Nie{\ss}ner, and
  Luisa Verdoliva.
\newblock Id-reveal: Identity-aware deepfake video detection.
\newblock In {\em Proceedings of the IEEE/CVF International Conference on
  Computer Vision}, pages 15108--15117, 2021.

\bibitem{cozzolino2018forensictransfer}
Davide Cozzolino, Justus Thies, Andreas R{\"o}ssler, Christian Riess, Matthias
  Nie{\ss}ner, and Luisa Verdoliva.
\newblock Forensictransfer: Weakly-supervised domain adaptation for forgery
  detection.
\newblock {\em arXiv preprint arXiv:1812.02510}, 2018.

\bibitem{draelos2020use}
Rachel~Lea Draelos and Lawrence Carin.
\newblock Use hirescam instead of grad-cam for faithful explanations of
  convolutional neural networks.
\newblock {\em arXiv e-prints}, pages arXiv--2011, 2020.

\bibitem{du2019towards}
Mengnan Du, Shiva Pentyala, Yuening Li, and Xia Hu.
\newblock Towards generalizable forgery detection with locality-aware
  autoencoder.
\newblock 2019.

\bibitem{durall2019unmasking}
Ricard Durall, Margret Keuper, Franz-Josef Pfreundt, and Janis Keuper.
\newblock Unmasking deepfakes with simple features.
\newblock {\em arXiv preprint arXiv:1911.00686}, 2019.

\bibitem{fernandes2019predicting}
Steven Fernandes, Sunny Raj, Eddy Ortiz, Iustina Vintila, Margaret Salter,
  Gordana Urosevic, and Sumit Jha.
\newblock Predicting heart rate variations of deepfake videos using neural ode.
\newblock In {\em Proceedings of the IEEE/CVF international conference on
  computer vision workshops}, pages 0--0, 2019.

\bibitem{fu2020axiom}
Ruigang Fu, Qingyong Hu, Xiaohu Dong, Yulan Guo, Yinghui Gao, and Biao Li.
\newblock Axiom-based grad-cam: Towards accurate visualization and explanation
  of cnns.
\newblock {\em arXiv preprint arXiv:2008.02312}, 2020.

\bibitem{haliassos2021lips}
Alexandros Haliassos, Konstantinos Vougioukas, Stavros Petridis, and Maja
  Pantic.
\newblock Lips don't lie: A generalisable and robust approach to face forgery
  detection.
\newblock In {\em Proceedings of the IEEE/CVF Conference on Computer Vision and
  Pattern Recognition}, pages 5039--5049, 2021.

\bibitem{hsu2020deep}
Chih-Chung Hsu, Yi-Xiu Zhuang, and Chia-Yen Lee.
\newblock Deep fake image detection based on pairwise learning.
\newblock {\em Applied Sciences}, 10(1):370, 2020.

\bibitem{jiang2021layercam}
Peng-Tao Jiang, Chang-Bin Zhang, Qibin Hou, Ming-Ming Cheng, and Yunchao Wei.
\newblock Layercam: Exploring hierarchical class activation maps for
  localization.
\newblock {\em IEEE Transactions on Image Processing}, 30:5875--5888, 2021.

\bibitem{khalid2021fakeavceleb}
Hasam Khalid, Shahroz Tariq, Minha Kim, and Simon~S Woo.
\newblock Fakeavceleb: a novel audio-video multimodal deepfake dataset.
\newblock {\em arXiv preprint arXiv:2108.05080}, 2021.

\bibitem{kim2021fretal}
Minha Kim, Shahroz Tariq, and Simon~S Woo.
\newblock Fretal: Generalizing deepfake detection using knowledge distillation
  and representation learning.
\newblock In {\em Proceedings of the IEEE/CVF Conference on Computer Vision and
  Pattern Recognition}, pages 1001--1012, 2021.

\bibitem{kwon2021kodf}
Patrick Kwon, Jaeseong You, Gyuhyeon Nam, Sungwoo Park, and Gyeongsu Chae.
\newblock Kodf: A large-scale korean deepfake detection dataset.
\newblock In {\em Proceedings of the IEEE/CVF International Conference on
  Computer Vision}, pages 10744--10753, 2021.

\bibitem{lee2021tar}
Sangyup Lee, Shahroz Tariq, Junyaup Kim, and Simon~S Woo.
\newblock Tar: Generalized forensic framework to detect deepfakes using weakly
  supervised learning.
\newblock In {\em IFIP International Conference on ICT Systems Security and
  Privacy Protection}, pages 351--366. Springer, 2021.

\bibitem{li2020face}
Lingzhi Li, Jianmin Bao, Ting Zhang, Hao Yang, Dong Chen, Fang Wen, and Baining
  Guo.
\newblock Face x-ray for more general face forgery detection.
\newblock In {\em Proceedings of the IEEE/CVF conference on computer vision and
  pattern recognition}, pages 5001--5010, 2020.

\bibitem{li2018ictu}
Yuezun Li, Ming-Ching Chang, and Siwei Lyu.
\newblock In ictu oculi: Exposing ai created fake videos by detecting eye
  blinking.
\newblock In {\em 2018 IEEE International Workshop on Information Forensics and
  Security (WIFS)}, pages 1--7. IEEE, 2018.

\bibitem{li2018exposing}
Yuezun Li and Siwei Lyu.
\newblock Exposing deepfake videos by detecting face warping artifacts.
\newblock {\em arXiv preprint arXiv:1811.00656}, 2018.

\bibitem{li2020celeb}
Yuezun Li, Xin Yang, Pu Sun, Honggang Qi, and Siwei Lyu.
\newblock Celeb-df: A large-scale challenging dataset for deepfake forensics.
\newblock In {\em Proceedings of the IEEE/CVF Conference on Computer Vision and
  Pattern Recognition}, pages 3207--3216, 2020.

\bibitem{marra2019gans}
Francesco Marra, Diego Gragnaniello, Luisa Verdoliva, and Giovanni Poggi.
\newblock Do gans leave artificial fingerprints?
\newblock In {\em 2019 IEEE Conference on Multimedia Information Processing and
  Retrieval (MIPR)}, pages 506--511. IEEE, 2019.

\bibitem{matern2019exploiting}
Falko Matern, Christian Riess, and Marc Stamminger.
\newblock Exploiting visual artifacts to expose deepfakes and face
  manipulations.
\newblock In {\em 2019 IEEE Winter Applications of Computer Vision Workshops
  (WACVW)}, pages 83--92. IEEE, 2019.

\bibitem{nguyen2019multi}
Huy~H Nguyen, Fuming Fang, Junichi Yamagishi, and Isao Echizen.
\newblock Multi-task learning for detecting and segmenting manipulated facial
  images and videos.
\newblock {\em arXiv preprint arXiv:1906.06876}, 2019.

\bibitem{nguyen2019use}
Huy~H Nguyen, Junichi Yamagishi, and Isao Echizen.
\newblock Use of a capsule network to detect fake images and videos.
\newblock {\em arXiv preprint arXiv:1910.12467}, 2019.

\bibitem{qian2020thinking}
Yuyang Qian, Guojun Yin, Lu Sheng, Zixuan Chen, and Jing Shao.
\newblock Thinking in frequency: Face forgery detection by mining
  frequency-aware clues.
\newblock In {\em European Conference on Computer Vision}, pages 86--103.
  Springer, 2020.

\bibitem{rossler2019faceforensics++}
Andreas Rossler, Davide Cozzolino, Luisa Verdoliva, Christian Riess, Justus
  Thies, and Matthias Nie{\ss}ner.
\newblock Faceforensics++: Learning to detect manipulated facial images.
\newblock In {\em Proceedings of the IEEE/CVF International Conference on
  Computer Vision}, pages 1--11, 2019.

\bibitem{selvaraju2017grad}
Ramprasaath~R Selvaraju, Michael Cogswell, Abhishek Das, Ramakrishna Vedantam,
  Devi Parikh, and Dhruv Batra.
\newblock Grad-cam: Visual explanations from deep networks via gradient-based
  localization.
\newblock In {\em Proceedings of the IEEE international conference on computer
  vision}, pages 618--626, 2017.

\bibitem{sjalander+:2019epic}
Magnus Sj\"alander, Magnus Jahre, Gunnar Tufte, and Nico Reissmann.
\newblock {EPIC}: An energy-efficient, high-performance {GPGPU} computing
  research infrastructure, 2019.

\bibitem{sun2022dual}
Ke Sun, Taiping Yao, Shen Chen, Shouhong Ding, Jilin Li, and Rongrong Ji.
\newblock Dual contrastive learning for general face forgery detection.
\newblock In {\em Proceedings of the AAAI Conference on Artificial
  Intelligence}, volume~36, pages 2316--2324, 2022.

\bibitem{tan2019efficientnet}
Mingxing Tan and Quoc Le.
\newblock Efficientnet: Rethinking model scaling for convolutional neural
  networks.
\newblock In {\em International conference on machine learning}, pages
  6105--6114. PMLR, 2019.

\bibitem{thies2019deferred}
Justus Thies, Michael Zollh{\"o}fer, and Matthias Nie{\ss}ner.
\newblock Deferred neural rendering: Image synthesis using neural textures.
\newblock {\em ACM Transactions on Graphics (TOG)}, 38(4):1--12, 2019.

\bibitem{thies2016face2face}
Justus Thies, Michael Zollhofer, Marc Stamminger, Christian Theobalt, and
  Matthias Nie{\ss}ner.
\newblock Face2face: Real-time face capture and reenactment of rgb videos.
\newblock In {\em Proceedings of the IEEE conference on computer vision and
  pattern recognition}, pages 2387--2395, 2016.

\bibitem{van2008visualizing}
Laurens Van~der Maaten and Geoffrey Hinton.
\newblock Visualizing data using t-sne.
\newblock {\em Journal of machine learning research}, 9(11), 2008.

\bibitem{wang2022m2tr}
Junke Wang, Zuxuan Wu, Wenhao Ouyang, Xintong Han, Jingjing Chen, Yu-Gang
  Jiang, and Ser-Nam Li.
\newblock M2tr: Multi-modal multi-scale transformers for deepfake detection.
\newblock In {\em Proceedings of the 2022 International Conference on
  Multimedia Retrieval}, pages 615--623, 2022.

\bibitem{xiao2015application}
Tianjun Xiao, Yichong Xu, Kuiyuan Yang, Jiaxing Zhang, Yuxin Peng, and Zheng
  Zhang.
\newblock The application of two-level attention models in deep convolutional
  neural network for fine-grained image classification.
\newblock In {\em Proceedings of the IEEE conference on computer vision and
  pattern recognition}, pages 842--850, 2015.

\bibitem{xu2022supervised}
Ying Xu, Kiran Raja, and Marius Pedersen.
\newblock Supervised contrastive learning for generalizable and explainable
  deepfakes detection.
\newblock In {\em Proceedings of the IEEE/CVF Winter Conference on Applications
  of Computer Vision}, pages 379--389, 2022.

\bibitem{yang2019exposing}
Xin Yang, Yuezun Li, and Siwei Lyu.
\newblock Exposing deep fakes using inconsistent head poses.
\newblock In {\em ICASSP 2019-2019 IEEE International Conference on Acoustics,
  Speech and Signal Processing (ICASSP)}, pages 8261--8265. IEEE, 2019.

\bibitem{zhang2016joint}
Kaipeng Zhang, Zhanpeng Zhang, Zhifeng Li, and Yu Qiao.
\newblock Joint face detection and alignment using multitask cascaded
  convolutional networks.
\newblock {\em IEEE Signal Processing Letters}, 23(10):1499--1503, 2016.

\bibitem{zhao2021multi}
Hanqing Zhao, Wenbo Zhou, Dongdong Chen, Tianyi Wei, Weiming Zhang, and Nenghai
  Yu.
\newblock Multi-attentional deepfake detection.
\newblock In {\em Proceedings of the IEEE/CVF Conference on Computer Vision and
  Pattern Recognition}, pages 2185--2194, 2021.

\bibitem{zhao2021learning}
Tianchen Zhao, Xiang Xu, Mingze Xu, Hui Ding, Yuanjun Xiong, and Wei Xia.
\newblock Learning self-consistency for deepfake detection.
\newblock In {\em Proceedings of the IEEE/CVF International Conference on
  Computer Vision}, pages 15023--15033, 2021.

\bibitem{zheng2021exploring}
Yinglin Zheng, Jianmin Bao, Dong Chen, Ming Zeng, and Fang Wen.
\newblock Exploring temporal coherence for more general video face forgery
  detection.
\newblock In {\em Proceedings of the IEEE/CVF International Conference on
  Computer Vision}, pages 15044--15054, 2021.

\bibitem{zhou2021joint}
Yipin Zhou and Ser-Nam Lim.
\newblock Joint audio-visual deepfake detection.
\newblock In {\em Proceedings of the IEEE/CVF International Conference on
  Computer Vision}, pages 14800--14809, 2021.

\bibitem{zhuang2020learning}
Peiqin Zhuang, Yali Wang, and Yu Qiao.
\newblock Learning attentive pairwise interaction for fine-grained
  classification.
\newblock In {\em Proceedings of the AAAI Conference on Artificial
  Intelligence}, volume~34, pages 13130--13137, 2020.

\end{thebibliography}
}

\clearpage
\newpage
\onecolumn
\section*{Supplementary Material: Learning Pairwise Interaction for Generalizable DeepFake Detection}
\label{sec:appendix}
\subsection*{Magnified Figures For \cref{fig:activation-map}}
\begin{figure}[htp]
    \centering
    \subfigure[Visualization of the last block of the exit flow of MCX-API (RGB).]{\label{fig:visualization-MCXapi}\includegraphics[width=\linewidth]{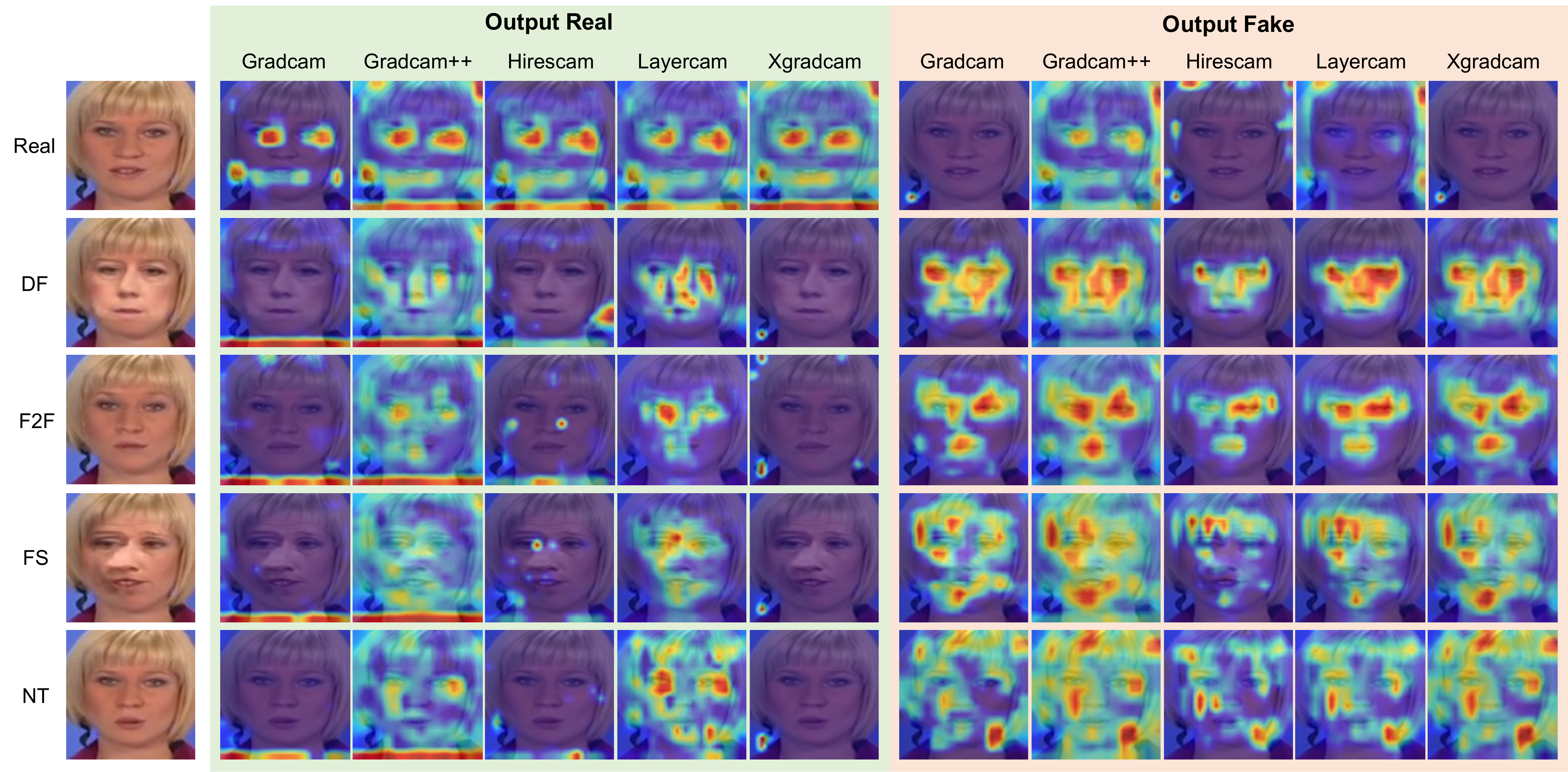}}
    \subfigure[Visualization of the last block of the API-Net.]{\label{fig:visualization-api}\includegraphics[width=\linewidth]{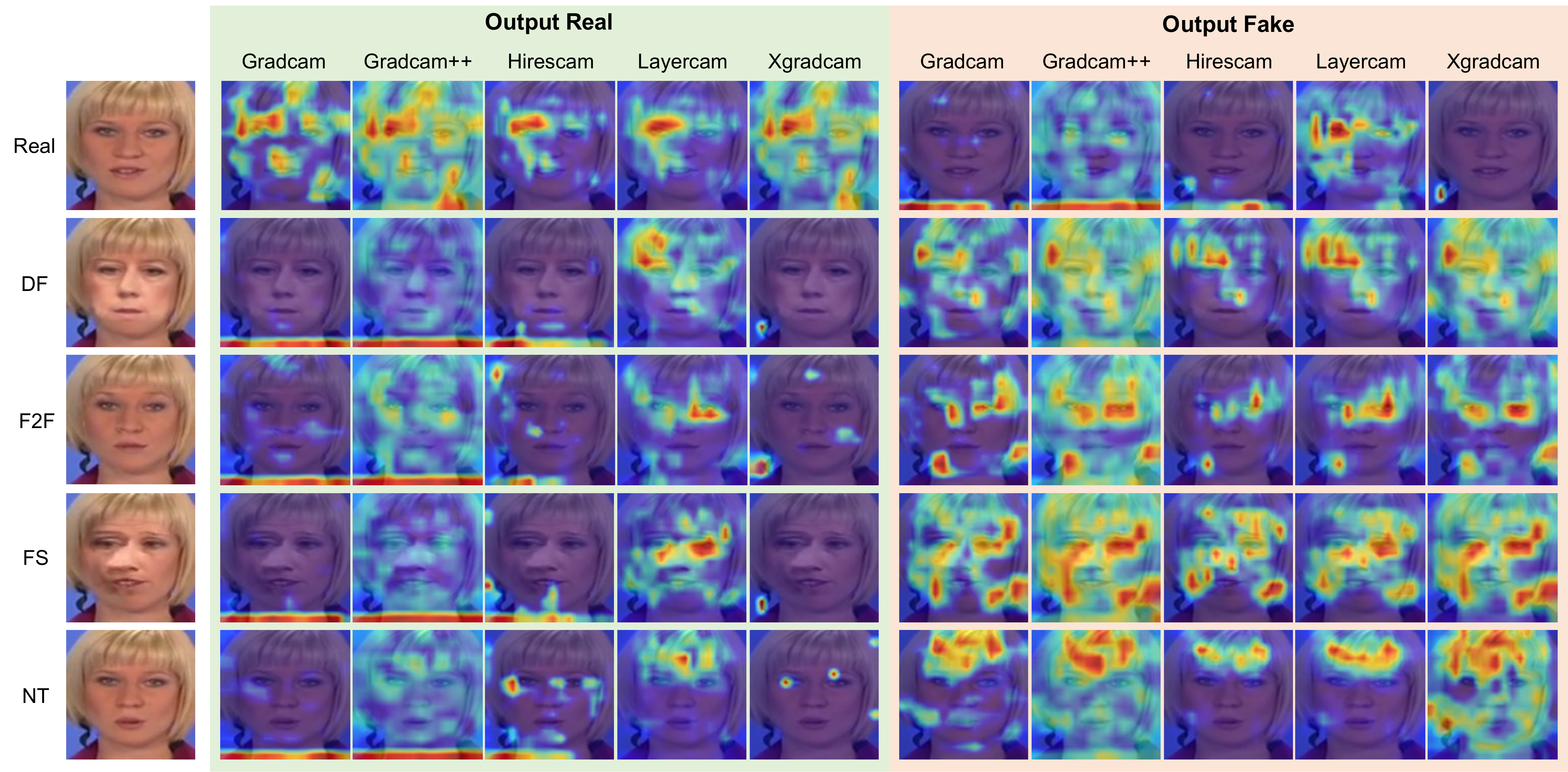}}
    
    \caption{A magified figure for \cref{fig:activation-map}. Visualization of the last layer of MCX-API (RGB) and API networks. We utilize Grad-CAM~\cite{selvaraju2017grad}, Grad-CAM++~\cite{chattopadhay2018grad}, HiResCAM~\cite{draelos2020use}, LayberCAM~\cite{jiang2021layercam} and XGradCAM~\cite{fu2020axiom} as our visualization tool.}
    \label{fig:activation-map-large}
\end{figure}
\end{document}